\documentclass[12pt]{article}

\usepackage[left=2.5cm,right=2.5cm,top=3cm,bottom=3cm]{geometry}
\usepackage{amsmath,amssymb}
\usepackage{longtable}
\usepackage{array}
\usepackage{fancyhdr}
\usepackage{titlesec}
\usepackage{tikz}
\usepackage[hidelinks]{hyperref}
\usepackage{enumitem}
\usepackage{multicol}
\usepackage{caption}
\usepackage{graphicx}
\usepackage{booktabs}
\newcolumntype{L}[1]{>{\raggedright\arraybackslash}p{#1}}

\title{Can the current trends of AI handle a full course of mathematics?}
\date{\today}
\author{
  Mariam Alsayyad\thanks{Faculty of Professional Studies, School of Foundation, Bahrain Polytechnic, Isa Town, Bahrain. \texttt{mariam.alsayyad@polytechnic.bh}}
  \and
  Fayadh Kadhem\thanks{Faculty of Professional Studies, School of Foundation, Bahrain Polytechnic, Isa Town, Bahrain. \texttt{fayadh.kadhem@polytechnic.bh}}
}

\begin{document}
\maketitle
\renewcommand{\thefootnote}{}%
\addtocounter{footnote}{-1}
\begin{abstract}
This paper addresses the question of how able the current trends of Artificial Intelligence (AI) are in managing to take the responsibility of a full course of mathematics at a college level. The study evaluates this ability in four significant aspects, namely, creating a course syllabus, presenting selected material, answering student questions, and creating an assessment. It shows that even though the AI is strong in some important parts like organization and accuracy, there are still some human aspects that are far away from the current abilities of AI. There is still a hidden emotional part, even in science, that cannot be fulfilled by the AI in its current state. This paper suggests some recommendations to integrate the human and AI potentials to create better outcomes in terms of reaching the target of creating a full course of mathematics, at a university level, as best as possible.
\end{abstract}

\section{Introduction}
Despite its invention before, back in late 2022, the revolution of AI amazed the world. A similar shock to that the people have faced when the search engines first rose in the late 1990s, the AI is doing the same now. Previously, people used to do their search in textbooks and libraries when they were looking for some information, but the search engines, like Google, Yahoo, and others, have amazingly shortened the process. Many questions and concerns arose at that time, and similarly, many have arisen now. It is widely agreed that constraints and restrictions must be put on AI to prevent it from being a serious cause and reason for reducing human ability, and to prevent facing ethical issues from relying heavily on it.\\

However, as happened with browsers and search engines, it is not wise to deal with the world now as it was before the AI. There is a powerful tool in our hands that can make a great contribution to life and provide many significant benefits. Despite the restrictions and constraints, our goal here is to test the ability of this tool in college life, especially in mathematics classes. The main goal is to see how and to what extent it is reliable and what can be done to integrate the specialist and AI strengths to come up with the best outcome for creating a full course in mathematics.\\

Our main approach was to do a work sample of a course starting from scratch and ask the AI to do a similar work in parallel. Then, three questionnaires have been made, which present the work of both without mentioning which is from humans and which is from AI. The questionnaires have then been distributed among experts, and their comments and observations have been studied by the authors to come up with the results of this paper.\\

It has been noted that there are indeed some important aspects that people can rely on AI for. In particular, the ability to organize information and integrate technology was a notable strength in all the AI work. This might make it more accessible and effectively accepted to students, as mentioned in \cite{SHSK}. However, as they also concluded, other factors and relationships are needed to be taken into account when AI is used in math education. Our main result is that the AI can do the work with the right orders coming from people, but the final version has to come from humans after revising the work and adding their emotions and feelings to it, even in the natural science fields.

\section{Literature Review}
 The emergence of AI has had an impact on everyone and almost every industry \cite{KK}. Basically, it applies massive technologies and algorithms and autonomously provides human-like intelligent solutions \cite{ZMBG}, with an extraordinary capacity to work through uncertain and complex situations \cite{GPR}. With the increasing popularity of the Internet of Things in education \cite{A}, the technology has been significantly incorporated into education \cite{KK}, and AI applications have significantly impacted teaching and learning strategies \cite{A}. \\
 
ChatGPT is a natural language processing system (NLP) and a tool from OpenAI \cite{O} that utilizes chats similar to those of humans \cite{LHY} to meet the needs of users \cite{GPR}. ChatGPT is configured to produce diverse results, including text, videos, and code \cite{SMNMV}. The responses ChatGPT produces are considered a turning point in the world \cite{SMNMV} and have been described as a ‘revolution in productivity’ \cite{W} because of its impressive, creative, and innovative implementations \cite{ILABBAAABB}.\\

With regard to education, ChatGPT applications have proved their capabilities and promising applications in the educational sector \cite{HT} and \cite{KSKBDFGHKMNPPPSSSK}. Recently, it has been proven that expanding the use of modern technologies in education has the potential to strengthen the educational system and provide greater opportunities for advancement and development \cite{A}, in addition to its significant impact on alignment with global developments and shaping the future of learners \cite{M}.\\

\noindent \textbf{Application in Education}\\
 The application of AI in higher education is estimated to grow rapidly \cite{CC}. It has been incorporated to facilitate not only teaching and learning but also administrative work \cite{GPR}. These applications include, but are not limited to, analyzing the quality of education \cite{LWXL}, resource management in higher education \cite{CB}, generating personalized learning to meet individual student needs \cite{KK2, OVG, VAP, XY}, forecasting academic performance \cite{CSR}, integrating AI into the course curriculum \cite{HBN}, and studying the relationship between learners, educators, technology, and institutions \cite{BNA}. These applications are utilized by educators to provide solutions and are not accessible to learners \cite{SMGGRMBT}, to facilitate high-tech educational solutions and adaptive learning frameworks \cite{Z}. Universities that are adopting AI in both teaching and learning strategies, along with its administrative use, are indeed distinguishing themselves in the competitive world and aiming for strategic growth and development \cite{HL}.\\

\noindent \textbf{Application in Mathematics}\\
 Extant research on education and mathematics has studied the utilization of technology capabilities, the integration of modern teaching strategies, individualizing student needs, diversifying assessment tools, and forming effective teaching frameworks \cite{C}.\\
 
In the research conducted by \cite{C}, a questionnaire of 183 students over 3 years was conducted. The aim of this work was to answer the question, ‘How to make the development of mathematical skills faster in the age of smart technologies?’. The researchers aimed to determine successful learning approaches and evaluate the integration of AI tools to support motivational growth and long-term enhancement. The results show the positive impact of AI teaching practices, which, as they suggested, can supplement or substitute current learning approaches. It concluded that, in mathematics, using big data and AI would make mathematics more engaging and understandable in the future.\\

Meanwhile, the research conducted by \cite{WTAJ} included the perspective of educators, not only students, on the use of AI in teaching mathematics, specifically after the emergence of ChatGPT. Unlike \cite{C}, this work applies qualitative research methods, including interviews and examination of user experience, conducted in two stages. The results of this research show that the performance of ChatGPT relies on the complexity of the problems, the data input into the platform, and the details of the requirements provided by the user. They concluded that the integration of ChatGPT into mathematics education still requires cautious utilization.
The work of \cite{SMNMV} investigated the impact of ChatGPT on mathematics education, specifically for engineering students. The study examined the effect of using ChatGPT on students’ problem-solving skills, critical analysis, and group collaboration skills. The analysis revealed positive evaluations in the adaptation of AI tools and demonstrated strong confidence. It is worth mentioning that there was concern about the ability to obtain essential engineering skills.
\cite{SKGSN} conducted a research review covering 39 years (between 1975–2014) on the challenges of adapting learning approaches, specifically in mathematics. The research highlighted the benefits of utilizing technology in learning mathematics, including the promotion of effective learning outcomes and enhancement of students’ creative thinking capabilities.\\

Lastly, \cite{FPGSLPCB} focused on the capabilities of ChatGPT in solving mathematical problems by channeling data to ChatGPT and to other mathematically trained models used as reference models for comparison. The results showed that ChatGPT struggles to provide precise answers, even though it understands the question in some cases.\\

With that being addressed, and in addition to the accessibility of AI applications and advancements for educators, \cite{A} suggested the utilization of those tools starting from curriculum design, to implementation in teaching strategies, and assessments, aiming to achieve optimal teaching and learning outcomes.\\

\noindent \textbf{Design of Curricula}\\
 The work conducted by \cite{SHBA} applied quantitative research methods to collect data from faculty members in higher education, along with experts in curriculum design. The research results, obtained through correlation and regression analyses, showed that the utilization of AI in curriculum design has a strong positive correlation with student performance as influenced by adaptive AI learning strategies. Meanwhile, the regression analysis indicated that the AI-based curriculum demonstrated significant predictive power regarding students’ academic performance.\\
 
Moreover, the work by \cite{E} aimed to design an AI-based curriculum to benefit from technology integration, specifically the ability to personalize learning to address diverse learning preferences and requirements, and the use of individualized assessment tools to boost students’ performance and engagement.\\

Although the research by \cite{SMGGRMBT} addressed the gap between university graduates and workforce requirements, it was aimed at designing AI-based curricula capable of creating key development opportunities for students in all fields of study. They claimed that the global technological transformation is affecting governments and businesses; therefore, a revolution in education is essential. Furthermore, the research indicated that AI-based curricula benefit universities themselves in terms of quality improvement opportunities and achieving accreditations.
Although \cite{SLP} also linked curriculum design using AI to the job market, they approached it differently. This work aimed to use AI’s backpropagation method to identify the skills required in the job market and design higher education curricula based on those competencies.\\

Lastly, the work of \cite{SNC} focuses on the range of barriers and potential benefits of applying AI-based curricula in higher education. The challenges encountered by educators include limited skills and knowledge related to AI, a lack of self-confidence in using it, and, more specifically, insufficient knowledge in curriculum design itself.\\

\noindent \textbf{Assessment}\\
 The work presented in \cite{GDXT} examined the application of ChatGPT in assessment, specifically in higher education. The aim of the paper was to investigate the potential benefits of utilizing ChatGPT by educators to assess students’ critical thinking. The results showed that this tool offers various prospects, including the accuracy of generated answers, writing excellence, and precise outputs, especially for academic work. Hence, it is recommended by the authors that educational stakeholders integrate AI into higher education to benefit from numerous educational tools \cite{LKKKC}.
Researchers in \cite{OAIEB} examined AI-based assessments in various aspects, including question generation, writing exam instructions, post-test administration, grading, grade analysis, and test report generation. Moreover, the study showed that although AI-based assessments offer strategic advantages and can redefine educational experiences, they still pose risks.\\

The work presented in \cite{RTT} analyzed the potential benefits and risks of utilizing assessments generated by ChatGPT from multiple perspectives, including those of students, teachers, and the education system. Although, in many situations, ChatGPT’s outcomes were nonsensical, the researchers emphasized the extreme importance of utilizing AI tools to cope with technological advancements.\\

The research by \cite{HRDRH} tackled AI-based assessments differently. The study’s main objective was to describe the most frequently employed machine learning algorithms used to determine student success in assessments. The analysis provides educators with an analytical strategy to design assessments that are accurate and reliable.\\

\noindent \textbf{Responding to Questions}\\
 The work of \cite{TBTPBKSARB} assessed ChatGPT’s ability to respond to multiple fields of study, including education, management, law, and accounting. The research aimed to examine the relevance of its answers to higher education assessments. The results of this study showed that the strength of ChatGPT’s answers lay in the clarity and readability of its responses. However, the primary shortcoming was its failure to provide critical evaluations and arguments in an engaging manner. Moreover, it was concluded that ChatGPT should not be seen as a solution to all problems; rather, it should serve as a starting point for idea generation.

\section{Methodology}
In order to collect the information and the opinions, the authors distributed the work into three main questionnaires. Each of them represents two similar parallel works. The hidden human work was called A (Syllabus A, Presentation A, Answer A, Assessment A), while the AI work was called B. So, anytime this paper refers to some work by the letter A, it means that it was done by humans, and vice versa for B. It is worth mentioning here that none of the respondents was told which work was which.\\

\noindent \textbf{AI system choice}\\
\noindent Since ChatGPT is the most trending tool used currently in everyday life by people with different backgrounds, the authors have decided to choose it for this research. They are aware that there might be more developed or advanced tools like the ones used in \cite{HHHN} and \cite{SMGGRMBT}, but it is not the aim of the study to examine or develop such tools. Also, most of them are either private or academic prototypes that are not maintained after their
research project finishes \cite{VP}. Thus, throughout, unless otherwise stated, anytime the word ``AI" appears, it means that it is generated by ChatGPT, specifically GPT‑4o. It has come to our mind that there might be more trained ways or strategies to talk to GPT‑4o, but the choice of sticking with the normal use is based on the moderate natural use of instructors or students, not by AI experts.\\

The questionnaires have been answered by approximately 30 respondents, and the details are as follows:\\

\noindent \textbf{Syllabus and Course Presentation}\\
\noindent The first questionnaire consisted of two parts; Part 1 started with showing two full syllabi for a first mathematics course at a college level for an applied university, while Part 2 illustrated two presentations for the topic of ``limits" in calculus, one generated by the authors and one generated by the AI.\\

\noindent \textbf{Questions and Answers}\\
\noindent The second questionnaire presented 16 real questions that were previously asked to the authors in their careers in similar courses at the intended level. They wrote their answers and gave the AI the same questions exactly and asked it to provide answers. The detailed questions with answers of both humans and AI are represented below in the appendix.\\

\noindent \textbf{Assessment}\\
\noindent Lastly, the third questionnaire consisted of two midterm exams, which are called in this paper midterm assessments, that are aimed to be 90-minute exams. Both exams have a list of questions together with their grades, and they target the same material exactly.\\

\noindent \textbf{Respondents}\\
\noindent The respondents of the questionnaires were math college professors and instructors, math high school teachers, math graduate students, and some trusted professionals with closely related majors to mathematics, like physics or engineering. This questionnaire has been sent to friends outside, but most of the respondents were Bahraini citizens or residents.\\

\noindent \textbf{Quantitative-Qualitative Approaches}\\
\noindent To get the best analysis, each questionnaire consisted of a list of quantitative questions supported by another list of qualitative ones. Both types of questions received careful consideration and were analyzed thoroughly by the authors.

\section{Quantitative Results}
This section is divided into three subsections below.
\subsection{Syllabus and presentation comparison}

The first questionnaire was divided into two parts. The first part presented two syllabi: one generated by the authors and the other by AI. The second part of the questionnaire illustrated two presentations that were structured similarly. Each part asked several questions regarding what was shown. Below are the quantitative results of this questionnaire.\\

After reading both syllabi, 50\% of the respondents believed that the AI-generated syllabus better addressed the course aims for a first mathematics course in an applied university. The remaining 50\% were split between 36.7\% favoring the human-generated syllabus and 13.3\% considering both to be at the same level.\\

When asked which syllabus was more reasonable in terms of Course Learning Outcomes, the responses were distributed as follows: 60\% for AI, 30\% for human, and 10\% for equal.\\
Regarding the clarity of the course description, the AI syllabus received 46.7\% of the votes, the human one 33.3\%, and 20\% considered both to be at the same level.\\

Delving deeper, the human syllabus received an average rating of 3.43 out of 5 for its learning outcomes, while the AI-generated syllabus scored 3.80.\\
For assessment distribution, the human-generated syllabus received 3.47 out of 5, while the AI-generated one scored 4.10. As for the appropriateness of the types of assessments, the human syllabus scored 3.47 and the AI 3.90. Regarding the schedule of assessments, the scores were 3.60 for human and 3.80 for AI.\\
In terms of the weekly breakdown of material, the human syllabus received 3.83, slightly higher than the AI’s 3.70. However, the evaluation of how well the weekly breakdown aligned with the intended learning outcomes was 3.80 for the human syllabus and 3.97 for the AI. On the other hand, the evaluation of weekly workload was 3.87 for the human syllabus and 3.63 for the AI.\\

Overall, the logical sequence of the human syllabus was rated 4.10 out of 5, while the AI's received 3.70. However, the organization of the syllabi was evaluated at 3.67 for human and 3.93 for AI.\\

Moving to the second part of the questionnaire, both the authors and the AI presented a selected topic. The following results were obtained: In terms of clarity and ease of understanding, the human presentation received an average of 4.03 out of 5, compared to 3.70 for the AI. Regarding relevance, the scores were 4.27 for human and 4.00 for AI.\\
The human presentation scored 4.17 for clarity of objectives, while the AI received 3.93. The accuracy of mathematical content was rated 4.43 for humans and 4.23 for AI.\\

When respondents were asked about the presentation comprehensiveness and sufficiency, the human received an average of 3.87, while the AI's scored 3.47. The relevance of examples and exercises to the intended topic was rated 4.10 for human and 3.83 for AI. The flow and smoothness of the presentation were scored at 4.23 for humans and 3.73 for AI.\\

Overall, the success of the human presentation was evaluated at 4.07 out of 5, while the AI's one got 3.63 out of 5.\\

Interestingly, when considering both parts of the questionnaire, there is a clear preference for the AI in the fields related to organization, while the human one had a better appreciation in terms of the week-by-week plan. However, there is an obvious dominance of humans in the presentation section. This suggests that while AI has reached a high level in organizing information, it still lags behind specialists in delivering effective scientific presentations. 
%
%

\begin{longtable}{|p{9cm}|c|c|}
\hline
\textbf{Criterion} & \textbf{Human} & \textbf{AI} \\
\hline
\multicolumn{3}{|c|}{\textbf{Syllabus Evaluation}} \\
\hline
Better addressing course aims & 36.7\% & 50\% \\
\hline
More reasonable Course Learning Outcomes & 30\% & 60\% \\
\hline
Clarity of course description & 33.3\% & 46.7\% \\
\hline
Learning Outcomes (out of 5) & 3.43 & 3.80 \\
\hline
Assessment distribution (out of 5) & 3.47 & 4.10 \\
Appropriateness of assessment types (out of 5) & 3.47 & 3.90 \\
Assessment schedule (out of 5) & 3.60 & 3.80 \\
\hline
Weekly breakdown of material (out of 5) & 3.83 & 3.70 \\
Alignment with CLOs (out of 5) & 3.80 & 3.97 \\
Weekly workload (out of 5) & 3.87 & 3.63 \\
\hline
Logical sequence (out of 5) & 4.10 & 3.70 \\
Organization (out of 5) & 3.67 & 3.93 \\
\hline
\multicolumn{3}{|c|}{\textbf{Presentation Evaluation (out of 5)}} \\
\hline
Clarity and ease of understanding & 4.03 & 3.70 \\
Relevance to topic & 4.27 & 4.00 \\
Clarity of objectives & 4.17 & 3.93 \\
Accuracy of mathematical content & 4.43 & 4.23 \\
Comprehensiveness and sufficiency & 3.87 & 3.47 \\
Relevance of examples and exercises & 4.10 & 3.83 \\
Flow and smoothness & 4.23 & 3.73 \\
Overall success of presentation & 4.07 & 3.63 \\
\hline
\end{longtable}
\subsection{Comparison of Human and AI Answers to Student Questions}
The distribution of the metrics of this questionnaire was as follows:\\
\noindent \textbf{16 questions}\\
\textbf{12: preference for human}\\
\textbf{1: preference for both equally}\\
\textbf{2: equal analysis}\\
\textbf{1: preference for AI}\\

\noindent \textbf{Human:} Highest $72.4\%$, lowest $17.9\%$\\
\textbf{AI:} Highest $50\%$, lowest $6.9\%$\\

Going into details, the second questionnaire presented 16 real questions that had been asked by students to the authors at various stages of their teaching careers. Each question had four answer options: Answer A, Answer B (which secretly represented human and AI answers, respectively), Equal Analysis, and Neither.\\

Out of the 16 questions, 12 responses favored the human-generated answers, while only one showed a preference for the AI-generated response. The remaining three were either evenly split between human and AI or had the majority vote for Equal Analysis.

In terms of percentages, the highest preference score for a human answer was $72.4\%$, while the highest for an AI answer was $50\%$.

These results support the qualitative feedback from respondents and clearly reflect their preference for human-generated answers. It is evident that most believed the human responses were significantly better than those produced by AI.\\

The reader who is interested in more details regarding the student questions and the answers of the authors and the AI, together with the percentages, is referred to the appendix at the end of this paper.

\subsection{Assessment Comparison}

In terms of clarity, both assessments received one vote indicating a lack of clarity, while the remaining responses were divided between "very clear" and "moderately clear" for both. However, instructors preferred the clarity of the AI-generated exam more, with $81.5\%$ rating it as "very clear," compared to $55.6\%$ for the human-generated exam.

Regarding the level of difficulty, one-third of respondents felt the human-generated exam was hard, while $63\%$ rated it as moderate. In contrast, $37\%$ considered the AI-generated exam easy, and $55.6\%$ rated its difficulty as moderate.

More than three-quarters of respondents believed the number of questions in the human-generated version was sufficient and reasonable. With a slight decrease in percentage, about three-quarters felt the same about the AI-generated version.

As for grade distribution, about $52\%$ of respondents said the human exam was fair, and $44.4\%$ considered it moderately fair. For the AI-generated exam, the same categories received $55.6\%$ and $22.2\%$ respectively. It is worth mentioning here that $11.1\%$ of responses rated the AI exam’s grade distribution as unfair, while no such vote was recorded for the human-generated version.

In terms of relevancy to the intended learning outcomes, the percentages were close. Specifically, $59.3\%$ and $33.3\%$ rated the human-generated exam as relevant and highly relevant, respectively, while $55.6\%$ and $29.3\%$ did so for the AI-generated exam, indicating a slight preference for the human version. Furthermore, when explicitly asked which exam best covered the proposed topics, the responses were evenly split: $44.4\%$ for the human exam, $44.4\%$ for the AI, and $11.1\%$ indicating both were equal.

Finally, when asked which assessment they would choose as instructors for their students, out of the responses, $54.55\%$ preferred the human-generated exam and $45.45\%$ preferred the AI-generated one. This indicates a nearly equal preference for both.

\section{Qualitative Results}
A similar division to the previous section is considered here.
\subsection{Syllabus and presentation comparison}

\noindent \textbf{The Course Structure}\\
The analysis showed that the following were commonly shared between human and AI-generated syllabi. The course structure, starting from the course objectives, course pre-requisites, course weekly breakdown, assessment distribution, grade distribution, main and additional resources, textbooks, and finally the teaching and learning hours, were all indicated as strength points in both syllabuses. The structure was inherently covering the requirements of first-year college students as it provides a strong foundation and broad coverage of the knowledge required, taking into account the level of the students. Moreover, the clarity of the course structure, along with the detailed course description, provides learners with an easier and more logical path to follow in order to understand what they need to do and achieve.

However, looking at the course content itself, there was a clear difference between the two of them. The course content was one of the main strength points of the human syllabus. More specifically, it was claimed that it showed experience in generating it. The course content was comprehensive, detailed, and specific. Moreover, the course follows the standard mathematical content for first-year college students and utilizes standard textbooks. The course weekly breakdown follows the path of the textbook, making it more logical and chronologically ordered, especially that this course prepares them for other mathematics courses. Therefore, this course is inherently interconnected with other courses. Hence, the syllabus provides an extensive foundation for students. Additionally, the research and applied projects in this syllabus promote problem-solving skills and links between theoretical knowledge of mathematics to real-life applications.

Meanwhile, the AI syllabus was strong in integrating the use of technology in the course weekly breakdown in order to diversify the teaching methods and to link teaching mathematics to real-world context. The AI suggested the use of technology tools like \texttt{MATLAB}, \texttt{GeoGebra}, and \texttt{Desmos} to enhance learning and create a collaborative learning environment.

Additionally, unlike the human-generated syllabus, which focused mainly on calculus, the AI course content was designed to cover a broader range of mathematical concepts and explores diverse topics like algebra, functions, trigonometry, calculus, vectors, and complex numbers. To build a strong foundation for students of different majors, it focuses on generality.\\

\noindent \textbf{Limitations of Human-made Syllabus}\\
In contrast to some respondents, the ``over standardization,'' as described by other respondents, is one of the main limitations raised in the human syllabus. They explained this as standardization that may hinder the teacher’s creativity and could lead to failing in addressing some student needs. In this course, the focus is on calculus topics, making it less applicable for a wide range of majors. Also, since the course covers in-depth calculus topics, those students who graduated from high school with majors like commercial, literary, and industrial streams will face some serious issues and may hinder their academic advancement. It assumes that the enrolled students have a strong foundation of pre-requisites.

Another point to focus on is the alignment between the course learning outcomes and the course content. The analysis revealed that there were imbalances between the weeks to teach differentiation that spans over 11 weeks, in comparison to only 2 weeks of integration, which contradicts the learning outcome on achieving comprehensive knowledge of integration, causing a potential misalignment between them. Moreover, another criticism of the learning outcome is that it could be further improved by using measurable terminologies to achieve them. This would help in connecting the course outcomes to the content and would reduce the vagueness.

The last weak point mentioned was the use of real-life applications in the weekly breakdown. Although the course utilizes research and applied projects, the weekly planned topics lack real-life applications, which is an essential point for an applied university. The respondents suggested using technological tools, like \texttt{MATLAB} and \texttt{GeoGebra}, to bridge this gap in the classes.\\

\noindent \textbf{Limitations of AI-generated Syllabus}\\
The main criticized point in the AI course’s weekly breakdown is the use of a wide range of topics. It is claimed that the course is too optimistic to cover a variety of topics while still maintaining an in-depth analysis of them. As a result, only superficial coverage per topic will be done. This, as described, is an overly ambitious syllabus, attempting to cover multiple mathematics disciplines that typically require separate courses. Moreover, the time to cover each topic is insufficient (2 to 3 weeks only), resulting in students facing cognitive overload. Additionally, the little focus on important topics in calculus results in not being able to prepare the students for more advanced courses, like for example differential equations in the coming courses, as they only covered shallow topics in this first-year, first-semester course.

Furthermore, since the topics are widely covered from multiple sectors and aspects, there is not a single standardized book for the students to follow. As a result, students might face an issue in tackling the topics in multiple referenced textbooks. Lastly, although the AI syllabus integrated the use of technology tools in mathematics learning, it was only encouraged to be used by the students and not explicitly graded. Therefore, student engagement may be superficial to those software tools.\\

\noindent \textbf{Preferences to Human or AI Syllabus}\\
The results showed an equal preference percentage between human and AI syllabi (48\% each), while only 1 respondent equally evaluated them.

The reason behind the selection of the human syllabus is that the standardized curriculum competes internationally and may assist students who have transferred from other universities and finished the course to continue their studies. Moreover, comprehensive calculus coverage ensures an adequate depth of knowledge per concept. Furthermore, the course content is appropriate for a first math course in an applied university, and the topics are clearly linked to textbook lessons. Lastly, the course provides the knowledge required to succeed in other advanced mathematics courses that rely heavily on differential and integral calculus concepts.

On the other hand, the preferences for the AI syllabus were due to the following reasons. The course learning outcome was well written and focused on the intended topics. Moreover, the course plan is well organized, and the assessment variety and methods are seen as an advantage. Furthermore, the coverage of essential topics like vectors, complex numbers, and applied trigonometry, as described, are indeed relevant to calculus, and tackle problems with strong emphasis on real-life applications and integration of computational tools, aligning well with the practical focus of applied institutions. Lastly, the suggested integrated technology tools such as \texttt{MATLAB}, \texttt{Desmos}, and \texttt{GeoGebra} are indeed considered as industry standards for simulation and modeling, which is aligned with the goal of applied universities.\\

\noindent Moving to the other part, the presentation comparison, the following points are noted:\\

\noindent \textbf{Strength of Human-made Presentation}\\

\noindent \textbf{Clarity of Presentation Slides} \\
The clarity of the presentation content is linked to two aspects. First, the clarity of the learning objectives in alignment with the specified topic. Second, the clarity of the material itself: the delivery of ideas was simple, direct, specific, focused on mathematical concepts, detailed, and precise. It was claimed that the clarity of the slides, along with the detailed step-by-step instructions, allows learners to learn even without the lecturer’s explanation.\\

\noindent \textbf{Presentation Content} \\
\noindent The material presented in the human-generated slides was coherent, comprehensive, effectively sequenced, and well communicated to the learners. One of the key strengths of this slide material is the gradual flow of ideas and the effective sequencing of their communication. This was reflected in a strong foundation built by introducing the concept, supported with supplemental visual representation and graphs, followed by worked examples that highlight common student mistakes, real-life applications, and a variety of dedicated exercises and problems. Additionally, the slides included detailed and precise solution steps, supporting the wide range of examples and different solution techniques.\\

\noindent \textbf{Strength of AI-generated Presentation}\\
\noindent \textbf{Content Organization} \\
The content was organized systematically, starting with clear, actionable, and precise lesson objectives. It then progressed to a strong introduction focusing on definitions, importance, and applications of the topic, providing students with motivation for the lesson. The slides then directly addressed the learning objectives in a logical flow, moving from theory to examples and finally to practical applications. The examples were broken down into clear and easy-to-follow steps, reducing the complexity of the problems and making them accessible to learners of varying levels. Mathematical accuracy was consistently maintained, with no ambiguities or unclear notation. Furthermore, the visual representation was clean, enhancing readability and reducing the cognitive load on students.\\

\noindent \textbf{Suggestions for Improvement for Human-made Presentation}\\
\noindent \textbf{Categorize Methods} \\
A key suggestion to improve the human-generated slides is to recategorize them by grouping examples and exercises per technique. Since the slides were comprehensive, reorganizing the content would help students highlight and clearly distinguish key ideas. It was also suggested to create a flow chart to show different situations and direct students to the appropriate solution techniques. This guidance would help students make proper decisions during assessments. Additionally, including a summary slide at the end could recap key concepts and reinforce decision-making strategies. This rearrangement can also support self-paced learning, particularly beneficial for students who missed lectures.\\

\noindent \textbf{Suggestions for Improvement for AI-generated Presentation}\\
\noindent \textbf{Add Visual Representation} \\
Respondents suggested using visual representation to further enhance student engagement and conceptual understanding, particularly for visual learners. This was especially recommended during the conceptual introduction of the lesson, as it would help first-year and first-semester students from diverse backgrounds grasp new concepts more easily.\\

\noindent \textbf{Use Variety of Examples} \\
It was noted that the AI-generated class presentation lacked sufficient exercises. It was highly recommended to include more practice examples to go beyond theoretical concepts. Additionally, more challenging and diverse exercises were recommended to address the needs of high-achieving students. The limited number of exercises could hinder students' ability to identify and apply the correct solution techniques. Lastly, more real-world applications, interactive elements, and practical activities were suggested to enhance the student learning experience and increase engagement.\\

\noindent \textbf{Missing and Detailed Steps} \\
The AI-generated content was criticized for not providing sufficient step-by-step explanations for each example. It was recommended to expand the level of detail and provide clear justifications for each step, to support self-paced learning—especially for students who miss classes. It was also suggested to include notes on common mistakes students might make when dealing with various ideas.\\

\noindent \textbf{The Preferences} \\
The analysis shows that preferences for human-made slides and AI-generated slides were 59.375\% and 31.25\%, respectively, with 9.375\% selecting both equally. This indicates that almost twice as many respondents preferred the human-generated class material over the AI's. Respondents highlighted the following key strengths of the human-generated slides:

\begin{itemize}
  \item The slides are more organized, with a clear and logical sequence of topics that contribute to achieving the lesson objectives.
  \item They are more effective in delivering information to students, though this also depends on the instructor’s skills.
  \item The slides are comprehensive, showing clear efforts to explain the lesson intuitively using diverse methods.
  \item They offer extensive practice examples and demonstrate a wide variety of algebraic techniques crucial for developing topic proficiency.
  \item The content is student-centered, engaging students from the start with contextual motivation, visual representations, and progressive examples, with an emphasis on clear justifications.
  \item The visual supports make them well-suited for introducing and reinforcing core ideas—especially important for first-time learners.
\end{itemize}

\noindent In contrast, the AI-generated presentation was chosen for the following attributes:

\begin{itemize}
  \item Clear and logical structure with a strong foundational introduction.
  \item A systematic approach to introducing concepts, immediately followed by well-explained, step-by-step examples.
  \item Explicit and precise learning objectives that provide a clear roadmap for both the instructor and students.
  \item Logical progression from theory to application, directly addressing learning objectives and helping students understand the “why” behind the topic’s concepts—significantly increasing engagement and motivation.
\end{itemize}

\subsection{Comparison of Human and AI Answers to Student Questions}
Here, the majority of respondents are willing to use the answers generated by humans, unlike AI responses, where the respondents mostly did not prefer them to respond to the student concerns. A detailed comparison is provided in the following points.\\

\noindent \textbf{The level of students}\\
 Human answers took into consideration the level of the students and thought of the reason behind the confusion raised by them. Hence, they better directed the answers to clear their doubts. Thus, it can be said that AI is lacking \textit{when it comes to
precise computation and advanced deductive reasoning} \cite{LZZL}.  Unlike human answers, the first issue with AI responses was clarity. The respondents claimed that these answers do not reveal the confusion of the students or provide sufficient information. Instead, they may create more questions because the answers are abstract and short. Moreover, they claim that the answers are not suitable for the targeted age group and level of the course. As this is an introductory mathematics course, more elaboration is needed, while those answers were for high achievers or specialists. However, upon improving, rephrasing, and expanding them, they could then be used in a classroom. \\

\noindent \textbf{Detailed answers}\\
Secondly worthening, most respondents appreciated the detailed explanation of the humans, claiming that the answers are thorough, comprehensive, and illustrative. Though there was an opposite opinion to this, in which some cases admitted not to use them as long answers are too boring to the students and could create some confusion as the answers were not direct to the point, or could have unnecessary information. Meanwhile, other respondents suggested that the clarity of the human answers is due to having thorough examples and intuition. In contrast to AI answers, many respondents agreed that AI answers are concise and direct to the point. However, a contrasting opinion was that those answers are convincing the students forcefully that those are definitions and pre-approved knowledge.
 
\subsection{Assessment Comparison}
The analysis suggested that the percentage preferences between the assessment generated by humans and the one generated by AI were $54.54\%$ and $45.46\%$, respectively, with a slight difference (differs by 2 respondents only). The following points provide details on the preferred focus points in the two assessments.\\

\noindent \textbf{Assessment structure}\\
The analysis showed that AI’s assessment is well aligned with the learning outcomes of the course. It is focused on covering the required topics comprehensively. Moreover, it showed that the questions were broken into steps, and the description of the questions was clear. Hence, it made it easier for the students to follow than the human assessment. Furthermore, the grade distribution of AI’s assessment was mostly indicated to be appropriate.\\

\noindent \textbf{Level of question difficulty}\\
The level of questions in the assessment that is generated by humans differentiated student levels; it assessed them with questions that are reasonable for first-year college students. It is claimed that it covered the examination of all concepts, tested student knowledge, and targeted more diverse competencies. Although responses to AI’s assessment also focused on the comprehensive coverage of topics and the idea variations, the level of questions was less challenging and easier for the students. Generally, the analysis shows that human assessment differentiated the assessment questions at the course and student level more than the AI’s one.  

\section{Analysis and Discussion}

\subsection{Course syllabus}
From the quantitative and qualitative results above, there are a couple of points that can be extracted from this study. The first is that although the organization of the syllabus and course generated by humans was appreciated, the AI organization was superior in most points. This shows that the AI has a strong ability to organize and can be relied on to do such work. This could be attributed to many factors, firstly, according to \cite{Z}, AI creates a learning experience that supports collaborative learning. Additionally, the research by \cite{SHBA} concluded that the use of AI in designing a curriculum for universities positively impacted learning strategies and dynamics.\\

\noindent The promotion of the interaction and collaboration was observed in the hands-on problem-solving group tutorials, peer quizzes, collaborative problem solving and presentations, and the integration of e-learning tools like MATLAB, GeoGebra, or Desmos for visualization and illustration.  The results of applying them would potentially promote the educational experience to expand beyond the classroom \cite{RW}, and develop student’s interactions with both tutors and technology \cite{BNA}. This is in alignment to the work proposed by \cite{Z}, which suggests that the AI could be used as a mediator between learners, instructors, and learning tools.\\

\noindent On the other hand, the superiority of the human syllabus was indicated on the weekly breakdown of material, the weekly workload, and the overall logical sequence of the syllabus. This shows that what is considered the core of most syllabi, namely, the weekly breakdown, is better chosen by humans. This might be due to the alignment with the global standards in most universities around the world.\\

\noindent Mixing the points above, it is straightforward to get the result that the use of the human weekly breakdown, which is compatible with the standard textbooks, followed by the organization of AI, will produce a better version of the syllabus.

\subsection{Class presentation}
According to the previous highlights, it was obvious that the evaluation of the human presentation exceeds the AI-generated one by far in all aspects.\\

\noindent In detail, the motivation, the mention of common mistakes, and the visual aid were notable to the respondents for the human presentation. Also, they liked the gradual flow of material, the clarity of examples, the variety of examples and exercises, and the comments on the step-by-step solution. Moreover, some respondents emphasized their impression of the detailed solutions, as they promote self-learning, especially for students who missed any class. On the other hand, despite the fact that the AI content organization was appreciated and the accuracy of mathematical concepts was maintained through the slides, there were some complaints about the lack of sufficient examples and problems in assessing the students' understanding. Moreover, the step-by-step solution could be further enhanced and written with more details.\\

\noindent As one can see above, it is worth emphasizing here that although the human presentation was evaluated better in all aspects, some aspects were closely evaluated against the human score, including the relevance of the topic to the course, the clarity of the objectives, and the accuracy of the mathematical concepts.\\

\noindent However, the presentation created by the authors took an average of 3 hours per day, and it was done in 4 days, while the change requests for the AI were done in counted seconds. As a result, for time-saving and a great outcome, it is wise to build the foundation using human ideas and ask the AI to generate the presentation based on that. This matches the educators' feedback in \cite{ODL} on the importance of getting the benefits from the AI. Perhaps the instructors could do a final modification again, if needed. For instance, the introduction and motivation of AI were strong and appreciated by the respondents, which agrees with the work of \cite{LCJDGQ, XCLTDC} which focuses on the significant contribution of AI to student’s motivation, enthusiasm to learn \cite{HLL},  and  learning engagement \cite{K}. Therefore, it could be integrated alongside the human slides to provide a strong topic delivery. Also, even though the accuracy was high, there is still a chance that we might get some errors from the AI. Hence, \textit{educators are increasingly committed to helping students develop the critical thinking skills necessary to detect and counter such errors} \cite{LZZL}.

\subsection{Question Responses}
As the results show, the respondents, in most questions, preferred the answers generated by humans. One key point is that the difference between the answers of humans and AI is that the human responses ignored the level of the students.  The AI answers were claimed to be for experts, specialists, or high achievers. In some cases, the answers were simply a definition or a proven rule. However, those are not suitable for year 1 students; therefore, educators are here to provide intuitive answers that are comprehensive and easily communicated. With that being addressed and by providing the multiple detailed orders to AI, indicating that the students are in semester 1 year 1, asking it to provide details, and illustrating them with real-life examples, AI's answers might be better communicated. This is due to the shown potential and notable accuracy of the answers of AI, as noted in the conclusion of \cite{CA} regarding its feedback.\\

\noindent Another key point is that, although the answers provided were scientific, the respondents observed a hidden emotional intelligence when the human responded to the questions; they showed empathy and compassion in clearing out the doubts of the students, and this was observed within the details provided, the type of examples, and the communication style. This point is compatible with the results obtained in \cite{GDXT}; they say that \textit{combining AI with human capabilities such as creativity, cognitive skills, and social-emotional skills would bring the best outcome}, which shows their appreciation for these human abilities. Hence, AI answers could be improved by requesting that students' emotions, age, and previous knowledge be considered, and a caring and compassionate communication tone should be used. Additionally, \cite{T} called educators in higher education to integrate empathy, sympathy, and compassion in teaching and learning strategies, a non-reachable point upon the utilization of AI to respond to the students \cite{RTT}.\\

\noindent Lastly, the human answers were distinguished by their detail and thorough explanation. This was observed in the rating for the questions with thorough information being high and far exceeding AI’s. Therefore, by providing comprehensive responses and expanding them with examples that are suitable for this age group of students, AI’s answers could be used in the classroom.\\

\noindent Unlike human answers, the answers provided by AI excluded in many cases the details, which is one of the key limitations in providing a comprehensive understanding to the students. Like [A6], who addressed the same issue by indicating that AI-generated answers jump from one idea to another, without fully providing details of each. As a result, the analysis lacks the critical explanation of the problem.\\

\noindent However, it is worth noting that the answers generated by AI could be used as a starting point to generate response to the students, however, to get an in-depth analysis on certain problem, a detailed and critical demonstration is required, clear communication skills to connect the information together, as well as the convincing argumental skills of human are required \cite{TBTPBKSARB}. \\

\noindent Moreover, as indicated in \cite{A}, the ease of getting the information from AI by just chatting with it, together with its current accuracy, and the suggested improvements can help students perform better and understand more, as mentioned in the results of both \cite{HHHN} and \cite{HRDRH}.\\

\noindent The results are aligned with the work of \cite{TBTPBKSARB}, which suggests that AI should be used as a tool to generate ideas and answers; however, they need further stimulation and an in-depth analysis.\\

\noindent Therefore, by providing comprehensive responses and expanding them with examples that are suitable for this age group of students, AI’s answers could be used in the classroom. 

\subsection{Assessments}
The analysis on the comparison between the assessment generated by human and by AI reveled many key points. First, the clarity of AI assessment was rated, on average, higher than that of human assessment. This is due to the clarity of the description of the questions, which made them easier to follow by the students.
Second, it is commonly observed that the level of human assessment was higher than that of AI. The human assessment differentiated the level of students, and provided a variety of problems ranging from easy, moderate, to critical thinking problems. Thus, this was more aligned to the course and student levels. In alignment with \cite{I}, instructors are experienced in designing effective assessments that are remarkable with their critical analysis, comprehensive evaluation, and higher thinking cognitive skills, in contrast to AI, more specifically ChatGPT. According to \cite{OAIEB}, AI test questions focus on the standard types trained by the algorithms who generates them, therefore, it does not account for the diverse levels of students, that are only perceived by instructors, nor on their backgrounds, it basically generates a one-size-fit for all students, as indicated in the respondent’s feedback.\\

\noindent The lack of human interaction is another limitation to the use of AI, especially in education, and specifically in generating the assessments. assessment level is detrimental in accordance with the observations, signals, hints, distinctions, and clues generated from the human interactions between instructors and students, which, in contrast, are missed in the AI assessment generation \cite{OAIEB}. 
Lastly, although most respondents evaluated the grading of the AI assessment to have better clarity, $11\%$ of the respondents rated the grade distribution for be unfair, which, as a result, means that instructors cannot completely rely on the AI grade distribution. 
With that, the instructors may design the assessment itself, indicating the learning objectives to be tested, the key context and questions, and the grade distribution.\\

\noindent However, the grade distribution should be specified by the instructors as well, as there are some biases in the grade distribution, as mentioned, which could lead to an unfair assessment structure. The issue with the use of AI is that the AI algorithms might be sometimes difficult to understand the reason behind the biases in the grade distribution \cite{OAIEB}. 
Upon doing so, instructors can then ask the AI to generate an assessment based on the details provided, to be able to use the AI’s competencies, such as clarity, idea variation, alignment with course learning outcomes, and time saving. Assessments generated by AI are certainly time savers, as they generate complex assessment that involves high skills and various ideas within a few seconds, which, in contrast, takes a large time on a human basis \cite{GDXT}.\\

The result of this analysis supports the work of \cite{Graide2022,G,S}, which agrees on utilizing AI as an assistant tool to support tutors in preparing their assessment, by combining instructors' skills, knowledge, experience, and emotional considerations with the AI’s technical competencies. 

\subsection{Conclusion and Recommendations}
Several differences could be considered when comparing human performance with AI. The first factor is the speed, as obviously, AI’s greatest advantage is the generation of outcomes in seconds. Second, its ability to make immediate improvements, modifications, and expansions.  Lastly, energy consumption, with AI being a machine, there is no consideration for this factor, unlike humans, who are presented with their brain usage. However, it is indicated that AI’s outcomes, specifically in the teaching and learning, is still facing an issue with considering student’s emotions, which are experienced by the instructors, therefore it is suggested that, regardless of the AI’s level of intelligence, being a machine with no emotional considerations to the students still advances humans over it.\\

The results of this paper showed that AI possesses strong capabilities in many aspects. Moreover, it can occasionally perform better than non-specialized humans or be equivalent to specialized ones. However, there is still a clear lack of human creative touch in some of its outputs. Our work above highlighted its strengths in organization and technology, and how accurate it is in providing correct answers most of the time. Our recommendations are to
\begin{enumerate}
    \item benefit from this advanced tool in education, as it makes work easier and faster;
    \item always provide it with creative ideas, as such ideas are not expected to emerge on their own without human input;
    \item use it to gain general knowledge or a big-picture understanding, but refer to trusted resources to ensure its output is accurate;
    \item always refine its output and double-check it each time.
\end{enumerate}

\section{Future work}
In the preparation of this work, the authors had several discussions about how this study could be approached. They chose to questionnaire instructors and specialists about the main aspects of this project. However, it is worth questionnaireing real students about their understanding to some aspects of the paper, like the presentation of the selected material generated by both humans and AI. This is an idea that is planned to be addressed in future work. Also, more points can be discussed regarding the comparison of the work of humans and AI in a math course. The comparison of how humans and AI grade exam papers will be tackled in an upcoming work, following this current paper.

\section*{Acknowledgements}
The authors thank the management of the Foundation School and their colleagues in the Mathematics Department at Bahrain Polytechnic for providing a nice work environment and for useful discussions during the preparation of this work. They also thank the specialists and friends outside Bahrain Polytechnic who participated in filling out the questionnaires and conducting inspiring discussions regarding this work.

\newpage
\section*{Appendix}
The following table represents the questions and answers of questionnaire 2 that was mentioned above. The complement of the sum of percentages for each question below was divided between the votes for ``equal analysis" and ``neither answer", where the first option was more or equal in all the times. Indeed, it is significantly higher most of the time.

\begin{enumerate}[leftmargin=0pt, label=\textbf{\arabic*.}, itemsep=1em]

\item \textbf{Why is $\infty - \infty$ and $\dfrac{\infty}{\infty}$ not equal to 0 and 1, respectively?}

\begin{tabular}{@{}L{0.47\textwidth} L{0.47\textwidth}@{}}
\textbf{Human's Answer} $(60.7\%)$ & \textbf{AI's Answer} $(32.1\%)$\\
\toprule
We can say that $\infty$ is a description more than a number. We use it to denote something that is uncontrollably big, but it is usually not a number. Now, we can write $\infty$ twice, but each $\infty$ has a different power or strength. For example:
    \begin{itemize}
        \item It is clear that the sequence: $1, 10, 100, 1000, ...$ is going up to infinity.
        \item It is clear also that the sequence $1,2,3,4,...$ keeps going up to infinity.
    \end{itemize}
However, one can clearly see that the first sequence is growing up much faster than the second one. Hence, the strength of the first infinity is more than that of the second one, in fact, much more.

This is an example where $\infty - \infty$ is still equal to $\infty$; to see this, just subtract each term in the second sequence from the one of the similar position in the first. However, if the first and second infinities are of the same power we will have that $\infty - \infty$ equals 0.

Thus, the problem is that we have a hidden strength when we write $\infty$ and this why $\infty - \infty$ can have different answers depending on the strength of the first and second infinities. The same idea exactly applies to the case $\frac{\infty}{\infty}$. & These are indeterminate forms. \(\infty - \infty\) and \(\dfrac{\infty}{\infty}\) can lead to different results depending on the expressions involved, so they are undefined without further analysis. \\
\bottomrule
\end{tabular}

\vspace{250pt}

\item \textbf{Why \(\infty^2 = \infty\)? Should not \(\infty^2\) be bigger?}

\begin{tabular}{@{}L{0.47\textwidth} L{0.47\textwidth}@{}}
\textbf{Human's Answer} $(37\%)$& \textbf{AI's Answer} $(33.3\%)$\\
\toprule
The symbol $\infty$ in this context represents a description, not a usual number. Thus, when we square something that is dramatically large, we still get something very very large and unbounded. So, we still use the same symbol to describe it regardless of the strength of the resulting infinity. & Infinity is a concept, not a number. \(\infty^2\) is still just infinity because we're talking about unbounded growth, not a precise value. \\
\bottomrule
\end{tabular}

\item \textbf{Why \(\dfrac{x^3}{x} = x^2\)? I feel they are the same as we have \(x\) above and below.}\\
\begin{tabular}{@{}L{0.47\textwidth} L{0.47\textwidth}@{}}
\textbf{Human's Answer} $(21.4\%)$& \textbf{AI's Answer} $(35.7\%)$\\
\toprule
The symbol \(x^3\) means we have 3 copies of \(x\). Thus,
$$\dfrac{x^3}{x}=\dfrac{x \cdot x \cdot x}{x}.$$
Now, we can cancel out one of the \(x\)'s on the top with the one on the bottom. This will give us two \(x\)'s on the top and this is represented by \(x^2$. & Because \(\dfrac{x^3}{x} = x^{3-1} = x^2\), provided \(x \neq 0\). You're dividing powers of the same base by subtracting exponents. \\
\bottomrule
\end{tabular}

\item \textbf{When do we use L'Hôpital's rule and when do we factorize?}\\
\begin{tabular}{@{}L{0.47\textwidth} L{0.47\textwidth}@{}}
\textbf{Human's Answer} $(35.7\%)$& \textbf{AI's Answer} $(35.7\%)$\\
\toprule
Firstly, we need to substitute the limit values to check if the situation is indeterminate, for example, an $\frac{\infty}{\infty}$, or a $\frac{0}{0}$. Usually, if we have polynomials that are easy to factorize, we will be able to cancel some common factors out of the expression in the numerator and denominator. However, if we have some other functions, like exponential functions, logs, or trig functions, it would be better to go with L'Hôpital's rule here. It is actually determined by the experience and easiness of each method. & Use L'Hôpital's Rule when limits result in indeterminate forms like \(\dfrac{0}{0}\) or \(\dfrac{\infty}{\infty}\). Factorizing is useful when algebraic simplification is more straightforward. \\
\bottomrule
\end{tabular}

\vspace{150pt}
\item \textbf{In solving a quadratic equation, should not the negative solution be excluded?}\\
\begin{tabular}{@{}L{0.47\textwidth} L{0.47\textwidth}@{}}
\textbf{Human's Answer} $(39.3\%)$& \textbf{AI's Answer} $(28.6\%)$\\
\toprule
The answer is yes and no! In fact, and this is a common case, if we have a real-life problem that only allows positive answers, we will reject the negative solution(s) of a quadratic equation or a polynomial equation. This for example happens if we deal with a distance problem or something similar. However, if there is no such restriction, nothing forces us to exclude the negative solutions. So, in some cases, we have to keep them. Moreover, we might have problems that only allow negative solutions; in such scenarios, the excluded solutions will be the positive ones. Hence, all of that relies on the problem itself and its requirements. & Mathematically, both solutions are valid. In real-world contexts (like distance), negative answers might be excluded due to practicality. \\
\bottomrule
\end{tabular}

\item \textbf{Why is the limit still the same even if we have a hole in the graph?}\\
\begin{tabular}{@{}L{0.47\textwidth} L{0.47\textwidth}@{}}
\textbf{Human's Answer} $(17.9\%)$& \textbf{AI's Answer} $(32.1\%)$\\
\toprule
The hole affects the value of a function at a specific point, but the limit studies the behavior of the function closer to the number from the positive and negative sides and does not focus on what is going on at the point itself. & Because a limit describes the behavior of a function as it approaches a value, not necessarily its value at that point. A hole doesn't affect the limit if both sides approach the same value. \\
\bottomrule
\end{tabular}
\newpage
\item \textbf{Why \((a + b)^2 \neq a^2 + b^2\)? What is the difference?}\\
\begin{tabular}{@{}L{0.47\textwidth} L{0.47\textwidth}@{}}
\textbf{Human's Answer} $(58.6\%)$& \textbf{AI's Answer} $(20.7\%)$\\
\toprule
First, notice that the LHS means that we multiply the expression $(a+b)$ by itself twice. As a result, this will give $a^2+2ab+b^2$, which is clearly not equal the second expression $a^2+b^2$. To see the difference clearly, one can pick some numbers for $a$ and $b$ and see the result of the LHS and RHS. For example, let us choose $a=1$ and $b=2$:
\[
\textnormal{LHS}=(a+b)^2=(1+2)^2=3^2=9
\]
\[\textnormal{RHS}=a^2+b^2=1^2+2^2=1+4=5\]
Clearly, the LHS and RHS are not equal here. So, this is why these expressions are not equal in general. & \((a + b)^2 = a^2 + 2ab + b^2\), not \(a^2 + b^2\). This is a common algebraic identity. \\
\bottomrule
\end{tabular}
%
\item \textbf{Why \(\sqrt{a + b} \neq \sqrt{a} + \sqrt{b}\)? What is the difference?}\\
\begin{tabular}{@{}L{0.47\textwidth} L{0.47\textwidth}@{}}
\textbf{Human's Answer} $(72.4\%)$& \textbf{AI's Answer} $(6.9\%)$\\
\toprule
It is not correct to assume that it is applicable to distribute the square root over the terms and add them to each other, as it is not distributive. The LHS means that we have to take the sum, then take the square root, while the second means we take the sum of the values after taking the square root. To see why these are not equal, we can see an example where they differ. A typical example is to take $a=9$ and $b=16$. In that case, we have:
\[
\textnormal{LHS}=\sqrt{a+b}=\sqrt{9+16}=\sqrt{25}=5\]
\[
\textnormal{RHS}=\sqrt{a}+\sqrt{b}=\sqrt{9}+\sqrt{16}=3+4=7
\]
So, it is clear that the equation is not true in general, since we have an example where the LHS and RHS give different answers. & The square root of a sum is not equal to the sum of square roots. \(\sqrt{a + b} \neq \sqrt{a} + \sqrt{b}\) unless one of them is 0. \\
\bottomrule
\end{tabular}
\item \textbf{Can an \(\infty\) be larger or smaller than other \(\infty\)? Basically, what is \(\infty\)?}\\
\begin{tabular}{@{}L{0.47\textwidth} L{0.47\textwidth}@{}}
\textbf{Human's Answer} $(44.8\%)$& \textbf{AI's Answer} $(34.5\%)$\\
\toprule
We can say that $\infty$ is a description more than a number. We use it to denote something that is uncontrollably big, but it is usually not a number. Now, we can write $\infty$ twice, but each $\infty$ has a different power or strength. For example:
    \begin{itemize}
        \item It is clear that the sequence: $1, 10, 100, 1000, ...$ is going up to infinity.
        \item It is clear also that the sequence $1,2,3,4,...$ keeps going up to infinity.
    \end{itemize}
However, one can clearly see that the first sequence is growing up much faster than the second one. Hence, the strength of the first infinity is more than that of the second one, in fact, much more. & All infinities are considered unbounded, but in advanced math like set theory or cardinality, different sizes of infinity exist (e.g., countable vs uncountable). \\
\bottomrule
\end{tabular}
\item \textbf{What is the difference between the equal (=) sign and the identity (\(\equiv\)) sign?}\\
\begin{tabular}{@{}L{0.47\textwidth} L{0.47\textwidth}@{}}
\textbf{Human's Answer} $(44.8\%)$& \textbf{AI's Answer} $(37.9\%)$\\
\toprule
First, we need to understand there are two types of equations, namely a conditional equation and an identity. The equal sign (\(=\)) is usually used to denote equations that are true for some specific values. For instance, one can write $\sin x =0$ and show that this is true when $x=n\pi$, where $n$ is an integer. However, the sign (\(\equiv\)) is used to indicate that the equation is generally true for all possible values. For instance, one can write $\sin^2x + \cos^2x \equiv 1$. Note that this is something symbolic, and actually there are some books that only use the equal sign (=) for both meanings. & \(=\) means equality for specific cases; \(\equiv\) means the two expressions are identical for all values (an identity). \\
\bottomrule
\end{tabular}
\newpage
\item \textbf{What is equating the coefficient? Is it a fixed rule?}\\
\begin{tabular}{@{}L{0.47\textwidth} L{0.47\textwidth}@{}}
\textbf{Human's Answer} $(55.2\%)$& \textbf{AI's Answer} $(20.7\%)$\\
\toprule
In general, two polynomials are equal if and only if
    \begin{itemize}
        \item their constant terms are equal,
        \item the factors in front of $x$ in both polynomials are equal,
        \item the factors in front of $x^2$ in both polynomials are equal,
        \item the factors in front of $x^3$ in both polynomials are equal, and so on.
    \end{itemize}
Hence, if we say, for example, that the polynomial $ax^4-x^2+bx-5$ is equal to the polynomial $9x^4+cx^3-dx^2+24x-5$, then each must equal its corresponding number of the other polynomial. Hence, in this example, we have: $a=9, b=24, d=1$ and $c=0$ (why?) This process is called equating coefficients, as we equate each letter with its correspondence from the other polynomial. & It's a method to compare coefficients of like terms on both sides of a polynomial equation. It's a standard algebraic approach. \\
\bottomrule
\end{tabular}
\newpage
\item \textbf{How is \(0!\) equal to 1?}\\
\begin{tabular}{@{}L{0.47\textwidth} L{0.47\textwidth}@{}}
\textbf{Human's Answer} $(65.9\%)$& \textbf{AI's Answer} $(10.3\%)$\\
\toprule
First, many textbooks say that $0!=1$, by definition. This means that they just force this equation to be true. However, if we want to see the intuition behind that, we can notice that, in general,
$$n!=\dfrac{(n+1)!}{n+1}.$$
Let us first verify this equation for some values of $n$. First, it is straightforward to see that $4!=24$ and $5!=120$ using the standard definition of factorial. Now, one can verify that the equation above is true for $n=4$ by doing the following:
$$4!=\dfrac{(4+1)!}{4+1}=\dfrac{5!}{5}=\dfrac{120}{5}=24.$$
Similarly, it is easy to see that $3!=6$, and it is easily verified that the equation above gives the following:
$$3!=\dfrac{(3+1)!}{3+1}=\dfrac{4!}{4}=\dfrac{24}{4}=6.$$
Hence, we can find the value of $0!$ using the same equation. This gives us:
$$0!=\dfrac{(0+1)!}{0+1}=\dfrac{1!}{1}=\dfrac{1}{1}=1.$$ & \(0! = 1\) by definition, and this keeps formulas (like combinations) consistent and meaningful. \\
\bottomrule
\end{tabular}
\vspace{250pt}
\item \textbf{Why is dividing any number by \(\infty\) equal to zero?}\\
\begin{tabular}{@{}L{0.47\textwidth} L{0.47\textwidth}@{}}
\textbf{Human's Answer} $(51.7\%)$& \textbf{AI's Answer} $(17.2\%)$\\
\toprule
First, let us do some calculations:
    \begin{itemize}
        \item $\dfrac{1}{2}=0.5$.
        \item $\dfrac{1}{10}=0.1$.
        \item $\dfrac{1}{100}=0.01$.
        \item $\dfrac{1}{1000}=0.001$.
        \item $\dfrac{1}{10000}=0.0001$.
    \end{itemize}
We can notice that when the number of the denominator becomes bigger, the result becomes smaller. Now, when we divide by $\infty$ this means we divide by the largest number that we can possibly imagine; hence, we should get the smallest nonnegative number we possibly imagine, which is 0. & As the denominator becomes infinitely large, the fraction approaches zero. It's a limit concept, not standard division. \\
\bottomrule
\end{tabular}

\item \textbf{What is the difference between factorizing \(a^3 + b^3\) and expanding \((a + b)^3\)?}\\
\begin{tabular}{@{}L{0.47\textwidth} L{0.47\textwidth}@{}}
\textbf{Human's Answer} $(62.1\%)$& \textbf{AI's Answer} $(20.7\%)$\\
\toprule
Factorization is used to simplify the expression as a product of its factors. Usually, it is used when other factors can be cancelled out from the expression. While expanding the polynomial is the opposite of the factorization, in which the factors are grouped together, rather than putting them in brackets. & Factorizing \(a^3 + b^3\) gives \((a + b)(a^2 - ab + b^2)\), a standard identity. Expanding \((a + b)^3\) gives \(a^3 + 3a^2b + 3ab^2 + b^3\), which is different. \\
\bottomrule
\end{tabular}
\vspace{250pt}
\item \textbf{What is the imaginary number \(j\) in a complex number?}\\
\begin{tabular}{@{}L{0.47\textwidth} L{0.47\textwidth}@{}}
\textbf{Human's Answer} $(57.1\%)$& \textbf{AI's Answer} $(21.4\%)$\\
\toprule
The imaginary number \(i\) or \(j\), depending on the book notation, is the square root of \(-1\), that is, \(\sqrt{-1}\). In fact, the square root is only defined on nonnegative numbers, but mathematicians have seen that many of the calculations made algebraically with the assumption that \(\sqrt{-1}\) has a value are correct. Moreover, it has been shown that many of the applications in physics and engineering are still correct with the same assumption. Hence, it has been agreed widely to assume that there is an \textit{imaginary} value for \(\sqrt{-1}\) and this is denoted by \(i\) or \(j\). A complex number is a mix of a real number and an imaginary number. For example, \(1 + j\), in which 1 is the real number, and \(j\) is the unit imaginary number. & In engineering, \(j\) is used instead of \(i\) to represent the square root of -1 to avoid confusion with electrical current. It is a complex unit. \\
\bottomrule
\end{tabular}
\newpage
\item \textbf{What is the difference between an expected event and the probability of an event?}\\
\begin{tabular}{@{}L{0.47\textwidth} L{0.47\textwidth}@{}}
\textbf{Human's Answer} $(42.9\%)$& \textbf{AI's Answer} $(50\%)$\\
\toprule
Probability is calculating the likelihood of a certain event to happen. The probability represents either a certainty that an event will occur, with a value of 1, or the impossibility that an event will occur with a value of 0, or a value that is between both 0 and 1. The probability is basically, the number of possibilities an event will occur, divided by the total number of possibilities. However, the expected outcome utilizes the probability of a series of events happened in the past, assuming that it will behave the same in the future. It is noteworthy to mention that the expected event could be of a value grater than 1, unlike the probability itself. In many cases, the expected outcome is represented similarly to the average value, as it averages the outcomes of a series of multiple past events. & The probability is the chance that an event will occur. An expected value is a weighted average of all possible outcomes based on their probabilities. \\
\bottomrule
\end{tabular}
\end{enumerate}

\begin{thebibliography}{99}

\bibitem{A}
Al Darayseh, A., 2023. \href{https://www.sciencedirect.com/science/article/pii/S2666920X23000115}{\textit{Acceptance of artificial intelligence in teaching science: Science teachers' perspective.}} Computers and Education: Artificial Intelligence, \textbf{4}, p.100132.

\bibitem{BNA}
Bearman, M., Nieminen, J.H. and Ajjawi, R., 2023. \href{https://doi.org/10.1080/02602938.2022.2069674}{\textit{Designing assessment in a digital world: An organising framework.}} Assessment $\&$ Evaluation in Higher Education, \textbf{48}(3), pp.291–304.

\bibitem{CSR}
Castrillón, O.D., Sarache, W. and Ruiz-Herrera, S., 2020. \href{http://doi.org/10.4067/S0718-50062020000100093}{\textit{Predicción del rendimiento académico por medio de técnicas de inteligencia artificial.}} Formación Universitaria, \textbf{13}, pp.93–102.

\bibitem{CB}
Chatterjee, S. and Bhattacharjee, K.K., 2020. \href{https://link.springer.com/article/10.1007/s10639-020-10159-7}{\textit{Adoption of artificial intelligence in higher education: A quantitative analysis using structural equation modelling.}} Education and Information Technologies, \textbf{25}, pp.3443–3463.

\bibitem{CC}
Chew, E. and Chua, X.N., 2020. \href{https://doi.org/10.1108/OTH-04-2020-0015}{\textit{Robotic Chinese language tutor: Personalising progress assessment and feedback or taking over your job?}} Horizont, \textbf{28}, pp.113–124.

\bibitem{CA}
Coskun, T.K. and Alper, A., 2024. \href{https://dialnet.unirioja.es/servlet/articulo?codigo=9624307}{\textit{{Evaluating the evaluators: A comparative study of AI and teacher assessments in Higher Education.}}} Digital Education Review, \textbf{(45)}, pp.124-140.

\bibitem{C}
Cunska, A., 2020. \href{https://www.europeanproceedings.com/article/10.15405/epiceepsy.20111.11}{\textit{Effective learning strategies and Artificial Intelligence (AI) support for accelerated math acquisition.}} In: European Proceedings of International Conference on Education and Educational Psychology. European Publisher.

\bibitem{FPGSLPCB}
Frieder, S., Pinchettil, L., Griffiths, R., Salvatori, T., Lukasiewicz, T., Petersen, P., Chevalier, A. and Berner, J., 2023. \href{https://doi.org/10.48550/arXiv.2301.13867}{\textit{Mathematical capabilities of ChatGPT.}} arXiv preprint arXiv:2301.13867.

\bibitem{E}
Ejjami, R., 2024. \href{https://jngr5.com/public/blog/The%20Future%20of%20Learning.pdf}{\textit{The future of learning: AI-based curriculum development.}} International Journal for Multidisciplinary Research, \textbf{6}(4), pp.1–31.


\bibitem{GDXT}
Gamage, K.A., Dehideniya, S.C., Xu, Z. and Tang, X., 2023. \href{https://journals.sfu.ca/jalt/index.php/jalt/article/view/1203/669}{\textit{ChatGPT and higher education assessments: More opportunities than concerns?.}} Journal of Applied Learning and Teaching, \textbf{6}(2), pp.358-369.

\bibitem{GPR}
González-Calatayud, V., Prendes-Espinosa, P. and Roig-Vila, R., 2021. \href{https://www.mdpi.com/2076-3417/11/12/5467}{\textit{Artificial intelligence for student assessment: A systematic review.}} Applied sciences, 11(12), p.5467.

\bibitem{Graide2022}
Graide, 2022. \textit{Graide}. \url{https://www.graide.co.uk/}

\bibitem{G}
Guilherme, A., 2019. \href{https://doi.org/10.1007/s00146-017-0693-8}{\textit{AI and education: The importance of teacher and student relations.}} AI \& SOCIETY, \textbf{34}(1), pp.47–54.

\bibitem{HL}
Hannan, E. and Liu, S., 2023. \href{https://doi.org/10.1108/CR-03-2021-0045}{\textit{AI: new source of competitiveness in higher education.}} Competitiveness Review: An International Business Journal, \textbf{33}(2), pp.265–279.

\bibitem {HHHN}
Henkel, O., Horne-Robinson, H., Kozhakhmetova, N. and Lee, A., 2024, July. \href{https://arxiv.org/pdf/2402.09809}{\textit{Effective and scalable math support: Experimental evidence on the impact of an AI-math tutor in Ghana.}} In International Conference on Artificial Intelligence in Education (pp. 373-381). Cham: Springer Nature Switzerland.

\bibitem{HT}
Holmes, W. and Tuomi, I., 2022. \href{https://doi.org/10.1111/ejed.12533}{\textit{State of the art and practice in AI in education.}} European Journal of Education, \textbf{57}(4), pp.542–570. 

\bibitem{HRDRH}
Hooda, M., Rana, C., Dahiya, O., Rizwan, A. and Hossain, M.S., 2022. \href{https://onlinelibrary.wiley.com/doi/full/10.1155/2022/5215722}{\textit{Artificial intelligence for assessment and feedback to enhance student success in higher education.}} Mathematical Problems in Engineering, 2022(1), p.5215722.

\bibitem{HBN}
Hornberger, M., Bewersdorff, A. and Nerdel, C., 2023. \href{https://www.sciencedirect.com/science/article/pii/S2666920X23000449}{\textit{What do university students know about Artificial Intelligence? Development and validation of an AI literacy test.}} Computers and Education: Artificial Intelligence, \textbf{5}, p.100165.

\bibitem{HLL}
Hou, J., Li, Z. and Liu, G., 2022. \href{https://doi.org/10.1155/2022/4295887}{\textit{Macro education approach to improve learning interest under the background of artificial intelligence.}} Wireless Communications and Mobile Computing, \textbf{2022}, Article 4295887.


\bibitem{ILABBAAABB}
Ibrahim, H., Liu, F., Asim, R., Battu, B., Benabderrahmane, S., Alhafni, B., Adnan, W., Alhanai, T., AlShebli, B., Baghdadi, R. and Bélanger, J.J., 2023. \href{https://www.nature.com/articles/s41598-023-38964-3.pdf}{\textit{Perception, performance, and detectability of conversational artificial intelligence across 32 university courses.}} Scientific reports, \textbf{13}(1), p.12187.

\bibitem{I}
Ifelebuegu, A., 2023. \href{https://doi.org/10.37074/jalt.2023.6.2.29}{\textit{Rethinking online assessment strategies: Authenticity versus AI chatbot intervention.}} Journal of Applied Learning and Teaching, \textbf{6}(2), pp.1–14.

\bibitem{K}
Karsenti, T., 2019. \href{https://doi.org/10.18162/fp.2019.a166}{\textit{Artificial intelligence in education: The urgent need to prepare teachers for tomorrow’s schools.}} Formation et Profession [Education and Profession], \textbf{27}(1), pp.112–116.

\bibitem{KSKBDFGHKMNPPPSSSK}
Kasneci, E., Sessler, K., Küchemann, S., Bannert, M., Dementieva, D., Fischer, F., Gasser, U., Groh, G., Günnemann, S., Hüllermeier, E., Krusche, S., Kutyniok, G., Michaeli, T., Nerdel, C., Pfeffer, J., Poquet, O., Sailer, M., Schmidt, A., Seidel, T. and Kasneci, G., 2023. \href{https://doi.org/10.1016/j.lindif.2023.102274}{\textit{ChatGPT for good? On opportunities and challenges of large language models for education.}} Learning and Individual Differences, 103, Article 102274. 

\bibitem{KK}
Kim, N.J. and Kim, M.K., 2022, March. \href{https://www.frontiersin.org/journals/education/articles/10.3389/feduc.2022.755914/full}{\textit{Teacher’s perceptions of using an artificial intelligence-based educational tool for scientific writing.}} In Frontiers in education (Vol. 7, p. 755914). Frontiers Media SA.

\bibitem{KK2}
Kim, W.-H. and Kim, J.-H., 2020. \href{https://ieeexplore.ieee.org/document/8985177}{\textit{Individualized AI tutor based on developmental learning networks.}} IEEE Access, \textbf{8}, pp.27927–27937.

\bibitem{LZZL}
Liang, S., Zhang, W., Zhong, T. and Liu, T., 2024. \href{https://www.arxiv.org/pdf/2412.16543}{\textit{Mathematics and machine creativity: A questionnaire on bridging mathematics with AI}}. arXiv preprint arXiv:2412.16543.

\bibitem{LKKKC}
Limna, P., Kraiwanit, T., Jangjarat, K., Klayklung, P. and Chocksathaporn, P., 2023. \href{https://doi.org/10.37074/jalt.2023.6.1.32}{\textit{The use of ChatGPT in the digital era: Perspectives on chatbot implementation.}} Journal of Applied Learning and Teaching, \textbf{6}(1), pp.64–74.

\bibitem{LCJDGQ}
Lin, P.Y., Chai, C.S., Jong, M.S.Y., Dai, Y., Guo, Y. and Qin, J., 2021. \href{https://doi.org/10.1016/j.caeai.2020.100006}{\textit{Modelling the structural relationship among primary students’ motivation to learn artificial intelligence.}} Computers and Education: Artificial Intelligence, \textbf{2}, p.100006.

\bibitem{LHY}
Lin, C., Huang, A. and Yang, S., 2023. \href{https://doi.org/10.3390/su15054012}{\textit{Review of AI-driven conversational chatbots implementation methodologies and challenges (1999–2022).}} Sustainability, \textbf{15}, p.4012.

\bibitem{LWXL}
Liu, M., Wang, Y., Xu, W. and Liu, L., 2017. \href{https://www.igi-global.com/gateway/article/169205}{\textit{Automated scoring of Chinese engineering students’ English essays.}} International Journal of Distance Education Technologies, \textbf{15}, pp.52–68.


\bibitem{M}
Mahmoud, A., 2020. \href{http://iafh.net/index.php/IJRES/article/view/240}{\textit{Artificial intelligence applications: An introduction to education development in the light of coronavirus pandemic COVID-19 challenges.}} International Journal of Research in Educational Sciences, \textbf{3}(4), pp.171–224.

\bibitem{RTT}
Rudolph, J., Tan, S. and Tan, S., 2023. \href{https://journals.sfu.ca/jalt/index.php/jalt/article/view/689}{ChatGPT: Bullshit spewer or the end of traditional assessments in higher education?.} Journal of applied learning and teaching, 6(1), pp.342-363.

\bibitem{OVG}
Ocaña-Fernández, Y., Valenzuela-Fernández, L.A. and Garro-Aburto, L.L., 2019. \href{http://doi.org/10.20511/pyr2019.v7n2.274}{\textit{Inteligencia artificial y sus implicaciones en la educación superior.}} Propósitos y Representaciones, \textbf{7}.

\bibitem{O}
OpenAI, 2025. \textit{ChatGPT} (July 2025 version) [online]. Available at: \url{https://chat.openai.com/} [Accessed 27 Jul. 2025].


\bibitem{ODL}
Otero, N., Druga, S. and Lan, A., 2024. \href{https://arxiv.org/pdf/2412.03765}{\textit{A Benchmark for Math Misconceptions: Bridging Gaps in Middle School Algebra with AI-Supported Instruction.}} arXiv preprint arXiv:2412.03765.

\bibitem{OAIEB}
Owan, V.J., Abang, K.B., Idika, D.O., Etta, E.O. and Bassey, B.A., 2023. \href{https://www.ejmste.com/download/exploring-the-potential-of-artificial-intelligence-tools-in-educational-measurement-and-assessment-13428.pdf}{\textit{Exploring the potential of artificial intelligence tools in educational measurement and assessment.}} EURASIA Journal of Mathematics, Science and Technology Education, 19(8), p.em2307.

\bibitem{RW}
Roll, I. and Wylie, R., 2016. \href{https://link.springer.com/article/10.1007/s40593-016-0110-3}{\textit{Evolution and revolution in artificial intelligence in education.}} International journal of artificial intelligence in education, \textbf{26}(2), pp.582-599.

\bibitem{SHBA}
Samee Ullah, R., Hashim, F., Bandeali, M.M. and Akbar, A., 2025. \href{https://thecrsss.com/index.php/Journal/article/view/684}{\textit{Artificial Intelligence in Curriculum Design: A Roadmap for Adaptive and Personalized Learning in Higher Education}}. The Critical Review of Social Sciences Studies, 3(3), pp.304–322.

\bibitem{SMNMV}
Sánchez-Ruiz, L.M., Moll-López, S., Nuñez-Pérez, A., Moraño-Fernández, J.A. and Vega-Fleitas, E., 2023. \href{https://www.mdpi.com/2076-3417/13/10/6039}{\textit{ChatGPT challenges blended learning methodologies in engineering education: A case study in mathematics.}} Applied Sciences, \textbf{13}(10), p.6039.


\bibitem{S}
Selwyn, N., 2019. \href{https://www.wiley.com/en-gb/Should+Robots+Replace+Teachers%3F%3A+AI+and+the+Future+of+Education-p-9781509528967}{\textit{Should robots replace teachers?: AI and the future of education.}} John Wiley \& Sons.

\bibitem{SHSK}
Setälä, M., Heilala, V., Sikström, P. and Kärkkäinen, T., 2025, March. \href{https://arxiv.org/pdf/2501.14779}{\textit{{The Use of Generative Artificial Intelligence for Upper Secondary Mathematics Education Through the Lens of Technology Acceptance}}}. In Proceedings of the 40th ACM/SIGAPP Symposium on Applied Computing (pp. 74-82).

\bibitem{SLP}
Somasundaram, M., Latha, P. and Pandian, S.S., 2020. \href{https://www.sciencedirect.com/science/article/pii/S1877050920313430}{\textit{Curriculum design using artificial intelligence (AI) back propagation method.}} Procedia Computer Science, \textbf{172}, pp.134–138.

\bibitem{SMGGRMBT}
Southworth, J., Migliaccio, K., Glover, J., Glover, J.N., Reed, D., McCarty, C., Brendemuhl, J. and Thomas, A., 2023. \href{https://www.sciencedirect.com/science/article/pii/S2666920X23000061}{\textit{Developing a model for AI Across the curriculum: Transforming the higher education landscape via innovation in AI literacy.}} Computers and Education: Artificial Intelligence, \textbf{4}, p.100127.

\bibitem{SNC}
Su, J., Ng, D.T.K. and Chu, S.K.W., 2023. \href{https://www.sciencedirect.com/science/article/pii/S2666920X23000036}{\textit{Artificial intelligence (AI) literacy in early childhood education: The challenges and opportunities.}} Computers and Education: Artificial Intelligence, \textbf{4}, p.100124.

\bibitem{SKGSN}
Szabo, Z.K., Körtesi, P., Guncaga, J., Szabo, D. and Neag, R., 2020. \href{http://doi.org/10.3390/su122310113}{\textit{Examples of problem-solving strategies in mathematics education supporting the sustainability of 21st-century skills.}} Sustainability, \textbf{12}, p.10113.

\bibitem{T}
Tan, E., 2022. \href{https://doi.org/10.37074/jalt.2022.5.2.ss1}{\textit{‘Heartware’ for the compassionate teacher: Humanizing the academy through mindsight, attentive love, and storytelling.}} Journal of Applied Learning $\&$ Teaching, \textbf{5}(2), pp.152–159.

\bibitem{TBTPBKSARB}
Tenakwah, E.S., Boadu, G., Tenakwah, E.J., Parzakonis, M., Brady, M., Kansiime, P., Said, S., Ayilu, R., Radavoi, C. and Berman, A., 2023. \href{https://www.researchsquare.com/article/rs-2968456/v2}{\textit{Generative AI and higher education assessments: A competency-based analysis}}. Research Square. Available at: https://doi.org/10.21203/rs.3.rs-2968456/v2 [Accessed 03 Aug, 2023].

\bibitem{VP}
Van Vaerenbergh, S. and Pérez-Suay, A., 2022. \href{https://arxiv.org/pdf/2107.06015}{\textit{A classification of artificial intelligence systems for mathematics education.}} In Mathematics education in the age of artificial intelligence: How artificial intelligence can serve mathematical human learning (pp. 89-106). Cham: Springer International Publishing.

\bibitem{VAP}
Villegas-Ch, W., Arias-Navarrete, A. and Palacios-Pacheco, X., 2020. \href{https://www.mdpi.com/2071-1050/12/4/1500}{\textit{Proposal of an architecture for the integration of a chatbot with artificial intelligence in a smart campus for the improvement of learning.}} Sustainability, \textbf{12}, p.1500.

\bibitem{WTAJ}
Wardat, Y., Tashtoush, M.A., AlAli, R. and Jarrah, A.M., 2023. \href{https://doi.org/10.29333/ejmste/13272}{\textit{ChatGPT: A revolutionary tool for teaching and learning mathematics.}} Eurasia Journal of Mathematics, Science and Technology Education, \textbf{19}(7), p.em2286.

\bibitem{W}
Watkins, M.D., 2022. ‘A revolution in productivity’: What ChatGPT could mean for business. [online] IMD. Available at: \href{https://www.imd.org/ibyimd/technology/a-revolution-in-productivity-what-chatgpt-could-mean-for-business/}{\textit{$<$https://www.imd.org/ibyimd/technology/a-revolution-in-productivity-what-chatgpt-could-mean-for-business/$>$}} [Accessed 27 Jul. 2025].

\bibitem{XCLTDC}
Xia, Q., Chiu, T.K.F., Lee, M., Temitayo, I., Dai, Y. and Chai, C.S., 2022. \href{https://doi.org/10.1016/j.compedu.2022.104582}{\textit{A self-determination theory design approach for inclusive and diverse artificial intelligence (AI) K–12 education.}} Computers \& Education, \textbf{189}, p.104582.

\bibitem{XY}
Xiao, M. and Yi, H., 2021. \href{https://doi.org/10.1002/cae.22235}{\textit{Building an efficient artificial intelligence model for personalized training in colleges and universities.}} Computer Applications in Engineering Education, \textbf{29}, pp.350–358.


\bibitem{ZMBG}
Zawacki-Richter, O., Marín, V.I., Bond, M. and Gouverneur, F., 2019. \href{https://link.springer.com/content/pdf/10.1186/s41239-019-0171-0.pdf}{\textit{Systematic review of research on artificial intelligence applications in higher education–where are the educators?.}} International journal of educational technology in higher education, \textbf{16}(1), pp.1-27.

\bibitem{Z}
Zhao, C., 2024 \href{https://jeti.thewsu.org/index.php/cieti/article/view/209}{\textit{AI-assisted assessment in higher education: A systematic review.}} Journal of Educational Technology and Innovation, Vol. 6 No. 4 (2024).

\end{thebibliography}
\end{document}